\documentclass[10pt,letterpaper]{article}
\usepackage{times}
\usepackage{epsfig}
\usepackage{graphicx, caption}
\graphicspath{ {./Figures/} }
\usepackage{amsmath}
\usepackage{amssymb}
\usepackage{tabularx}
\usepackage{array}
\usepackage[strings]{underscore}
\newcolumntype{R}[1]{>{\raggedleft\arraybackslash}p{#1}}
\usepackage{siunitx}
\usepackage{wrapfig}
\usepackage{amsmath}
\usepackage[dvipsnames]{xcolor}
\usepackage[section]{placeins}
\usepackage{authblk}
\usepackage[a4paper, total={6in, 8in}]{geometry}

\usepackage[misc]{ifsym}

\usepackage[
backend=biber,
style=nature,
]{biblatex}
\makeatletter
\AtBeginDocument{%
  \expandafter\renewcommand\expandafter\subsection\expandafter{%
    \expandafter\@fb@secFB\subsection
  }%
}
\makeatother

\usepackage[breaklinks=true,bookmarks=false]{hyperref}

\begin{document}
%%%%%%%%% TITLE
\title{\textbf{Stitching and dimensionality effects on large artificially generated volume datasets}}
\author[1]{Lucas von Chamier}
\author[2,3,4]{Jan Philipp Albrecht} 
\author[2,3,5]{Dagmar Kainmueller}
\affil[1]{GFZ Helmholtz-Zentrum für Geoforschung}
\affil[2]{Max Delbrück Center for Molecular Medicine in the Helmholtz Association}
\affil[3]{Helmholtz Imaging}
\affil[4]{Humboldt-Universität zu Berlin}
\affil[5]{University of Potsdam}

\maketitle

\abstract{{Generating large images via deep learning requires patching input data to accommodate hardware memory limitations, then assembling output patches, a process that can introduce stitching artifacts when neighboring patches do not align at borders. While these artifacts are known to affect segmentation tasks, their impact on generative models for style-transfer remains poorly understood.
We investigated three stitching approaches and two patch dimensionalities (2D vs 3D) using cycleGAN models trained on cryo-electron microscopy datasets. We evaluated both perceptual quality and performance on downstream mitochondria segmentation.
Our key findings reveal that: (1) FID scores fail to detect subtle stitching artifacts that significantly impact downstream segmentation performance, (2) 3D models with artifact-free stitching marginally outperform 2D models on downstream tasks, though the improvement barely justifies the computational cost, and (3) 2D models train more stably due to larger batch sizes. Additionally, we demonstrate that ensembling predictions from three orthogonal directions can improve low-quality volumes but provides no benefit for high-quality outputs.
These results demonstrate that maximizing generative model performance on large scientific datasets requires careful consideration and mitigation of stitching artifacts, and that perceptual metrics alone are insufficient for evaluating domain adaptation quality in biomedical imaging.}}

\section{Introduction}

Image generation is one of the most innovative capabilities in the toolbox of recent AI methods and has generated great interest in image-based research, from super-resolving images from low resolution  inputs\cite{ledig2017photo, saharia2022image, demiray2021d, jansche2025deep} to style-transfer tasks \cite{isola2017image, zhu2017unpaired, lauenburg2022instance, kieselmann2021cross, zhang2018translating, de2021residual} and even modeling de novo data from text or image prompts \cite{khader2023denoising, thambawita2022singan, eschweiler20213d}. However, image generation via deep neural networks using current hardware comes with inherent computational limits in terms of the size of the data which can be processed in a single forward pass. Most neural network methods for generative image tasks are benchmarked on datasets of images with dimensions between 64 to 1024 pixels, usually only in 1 to 3 channels and in 2d \cite{yu2014fine, yu2017semantic, ILSVRC15, liu2015faceattributes}, while data acquired and analysed by researchers are frequently significantly larger in terms of size as well as the number of dimensions and channels.
This means that scientists interested in using generative methods for images must often assemble images from individual patches which fit into the respective hardware the underlying networks are housed in, in a process we will refer to as 'stitching'. 

Stitching is a common routine for image-based research\cite{preibisch2009globally, bria2012terastitcher}, and requires the mitigation of so-called stitching artefacts \cite{reina2020systematic, rumberger2021shift, buglakova2025tiling}. These artefacts occur when the patches images are assembled from do not align well along their borders, leading to visible seams in the full image which are not features of the underlying specimens. There are various factors which can lead to stitching artefacts, such as independent normalisation of patches, bad registration of neighbouring patches and artefacts based on the processing of individual patches. In generative AI methods for image generation, the latter artefacts can easily occur since the underlying networks process the borders of individual patches independently from another and therefore do not always align pixels of neighbouring patches perfectly. 
Additionally, many vision-based networks use padded convolutions allowing patches to maintain their dimensions even in very deep neural networks \cite{he2016deep}, which facilitates learning of more complex features and promises better performance for various vision tasks. However, padding adds pixels to the borders of patches which are not inherent in the input data, and does so in each layer it is used in, leading to the edges of generated patches containing information from pixels which do not exist in the input. Patches created in this way often have obvious artefacts with their neighbours.

There are various strategies to mitigate these artefacts, the most common of which is to process patches from a full image using a stride which is smaller than the patch size, such that there exists an overlap between neighbouring patches \cite{possolo2021exact, wolny2020accurate, saalfeld2019computational}. When the full image is assembled, the overlapping regions can be averaged or cropped (or a combination of these) in a way which makes stitching artefacts nearly imperceptible, though does not fully remove them \cite{rumberger2021shift}.
Another method which can be used in convolutional neural networks are convolutions without border-padding (valid convolutions) \cite{ronneberger2015u, rumberger2021shift} which mitigate some of the effects of introducing pixels and improve the shift-equivariance properties of these neural networks \cite{kayhan2020translation}. However, it comes at the cost of network depth, since valid convolutions shrink the input patches by a minimum of two pixels for each strided convolution across the patch. It also is not capable of removing all border-artefacts as we showed in previous research \cite{rumberger2021shift}.

Another aspect researchers on 3D images need to decide is whether to assemble their images from 2d patches, slice-by-slice, or using 3-dimensional patches. The former is the most commonly used approach, primarily due to the additional computational cost of loading 3d patches and 3d models into GPU memory. However, it is known from comparisons between 2d and 3d networks, that segmentation performance often improves when using 3d patches on the same datasets \cite{cciccek20163d, reina2020systematic}, presumably due to the increased context available for the network when encoding an image region. However, in generative models, the aspect of patch dimensionality and its effect on the overall quality of the image or downstream tasks has hardly been explored, even though the effect could be similarly advantageous.

\section{Summary of Results}
The primary focus of this research is to explore how stitching in a style-transfer task affects the perceptual quality of the output and the performance of downstream tasks. We investigate these questions on the case of cryoEM tomography data, acting as an example for large 3d image datasets. The secondary focus is to investigate how the choice of tile-dimensionality affects the same metrics. Our secondary focus emerged due to the 3d dimensionality of the data we explored. We were interested if the additional context available in 3d tiles of the data would allow the models to generate images with better overall quality and better performance on downstream tasks. Below we summarize the key findings and outcomes of this study.

\subsection{Main findings}
\begin{itemize}
\item The perceptual metric Frechet inception distance (FID) \cite{heusel2017gans} is not sensitive enough to predict the presence of stitching artefacts in images.
\item Downstream tasks are affected by the presence of stitching artefacts in generated input data, even when they are subtle.
\item The performance of 2d and 3d models differs little for downstream tasks, with the best 3d models only marginally better than the best 2d models, 
\item 2d models trained with larger batch sizes train significantly more stably than 3d models with lower batch sizes.
\item 3d volumes assembled from 2d tiles with low perceptual quality often improve by the use of ensembling across different dimensions, but already high-quality volumes do not. To our knowledge, this was the first time ensembling from three directions was tested for a style transfer task in 3d.
\end{itemize}

\subsection{Best practices}
We find that the best output quality in the predictions is achieved by using a tile-and-stitch approach as first established by Rumberger et al. \cite{rumberger2021shift}. This means using valid convolutions and cropping of output patches according to the downsampling factor and the number of downsampling steps (see \ref{sec:methods}).
To ensure seamless stitching of large images it is necessary to use global normalisation statistics for all patches of the dataset. The statistics should be collected during training and frozen for inference, such that the same normalisation is applied to every patch on an image.

\subsection{Code}
Our code is available to the public here: \url{https://github.com/lucpaul/3D_cycleGAN} and is a revised version of the cycleGAN method \cite{zhu2017unpaired}. Our main improvements to the existing pipeline include:
\begin{itemize}
\item a choice between 2d and 3d training and inference
\item a choice between different styles of stitching for inference
\item batched inference and use of mixed precision for accelerated training and generation of large volumes
\item memory-efficient data-loading for arbitrarily sized datasets
\end{itemize}

\section{Datasets and Models}
\subsection{Datasets}

\subsubsection{MitoEM}
This dataset contains two cryoEM stacks, from a rat and a human cell, respectively \cite{wei2020mitoem}. The datasets were made isotropic, to a voxel-size of 30x30x30nm, using bicubic interpolation and averaging. For cycleGAN training, this dataset was used as is with 1040 slices for the human dataset and 1000 for the rat dataset, respectively. 
For the downstream task of segmentation via a Unet, the dataset was split according to the original instructions into a training stack containing 400 annotated slices and and a test stack with 100 annotated slices. The masks were binarized for the semantic segmentation task. For the training with a patch size of 190, the stacks were split into 32.400 and 45.864 in 2d patches and 216 and 384 3d patches of rat and human data, respectively.
\label{subsubsec:mitoem}

\subsubsection{openorganelle dataset} This dataset was assembled from data available through the janelia openorganelle database \cite{heinrich2021whole} from jurkat \cite{jrcjurkat-1} and HeLa cells \cite{Xu2020, jrcHela-2, jrcHela-3}. The stacks were made isotropic in xyz by bicubic interpolation to a voxel size of 10x10x10nm, and cropped to reduce the fraction of background voxels. This dataset was used prior to the main experiment to test hyperparameters which yielded strong FID scores for style transfer with cycleGAN. No annotations were used to evaluate downstreamn performance on this data. 
For the larger patch size of 190 pixels/voxels in 2d, 117.104 and 171.043 patches, and for 3d, 710 and 1068 patches were generated, for jurkat and HeLa data, respectively.
For the smaller patch size of 158 pixels/voxels in 2d, 167.636 and 227.527 and for 3d, 1263 and 2003 patches were generated, respectively.
\label{subsubsec:openorganelle}

\subsection{cycleGAN}
The backbones for the cycleGAN generators were UNets with 5 downsampling steps. As in the original publication downsampling was achieved by using convolutions with stride 2 instead of pooling layers \cite{zhu2017unpaired}. Small modifications were made to the generator architecture. Specifically, all convolutions in the UNet generators were given valid padding (i.e., no padding) during training. Padding was only added to the convolutions once training was complete if the model was tested for the case of overlapping tiles.
Furthermore, the instance normalisation layers used in the original cycleGAN code, instead of using running statistics during both, training and inference, must track the statistics only for training batches but freeze these during inference to ensure that patches are not individually normalised during inference which could lead to patch artefacts in the final image.
An additional structural similarity (SSIM) loss term was added to the cycle-consistency loss, with the aim to conserve structural image features \cite{jin2017cyclegan}. An SSIM loss with a small weight (0.01) was added also to the forward cycle in addition to the critic loss, with the same aim.
The cycleGAN models for the main experiment on the \textit{MitoNet} dataset were trained for 200 epochs on a single Nvidia RTX3090 GPU, with constant learning rate of 2e-4, with checkpoints saved every 5 epochs. 
The experiment on the \textit{openorganelle} dataset was trained for 100 epochs. When using cosine learning rate scheduling, the learning rate started at 2e-4 and had a minimum of 0.0, with a frequency of 20 epochs, giving 5 cycles for this setting.
No other modifications were made to the architecture or training pipelines.

When loading images into the model during training, we excluded patches which contained entirely or mostly background pixels. We did this by measuring the variance across pixels in each patch, and rejected patches below the 5th percentile across the full dataset. This was done to reduce the impact of patches with almost no relevant style features from the training datasets.

\subsection{UNet for downstream segmentation}
We used a simple 3D Unet \cite{wolny2020accurate} with three up-down blocks with a convolution before each downsampling step. The model was trained using a constant learning rate with a 'decay-on-plateau' learning rate policy with a patience of 5 epochs and a decay factor of 0.5, with a starting learning rate of 0.0004.
All Unet models were trained for 100 epochs or until the learning rate reached 1e-8. The dataset was split into patches of 140x140x140 voxels, and the minimal volume of each patch which had to contain a label had to be 2\%. For inference, the tile-and-stitch approach was used to guarantee no stitching artefacts are introduced, with output tiles cropped to 96x96x96 voxels in size. The models were trained on a single Nvidia RTX3090 GPU.

\clearpage

\section{Experiments}

\begin{figure*}[!ht]
\centering
\captionsetup{width=0.89\textwidth}
    \includegraphics[width=0.89\textwidth]{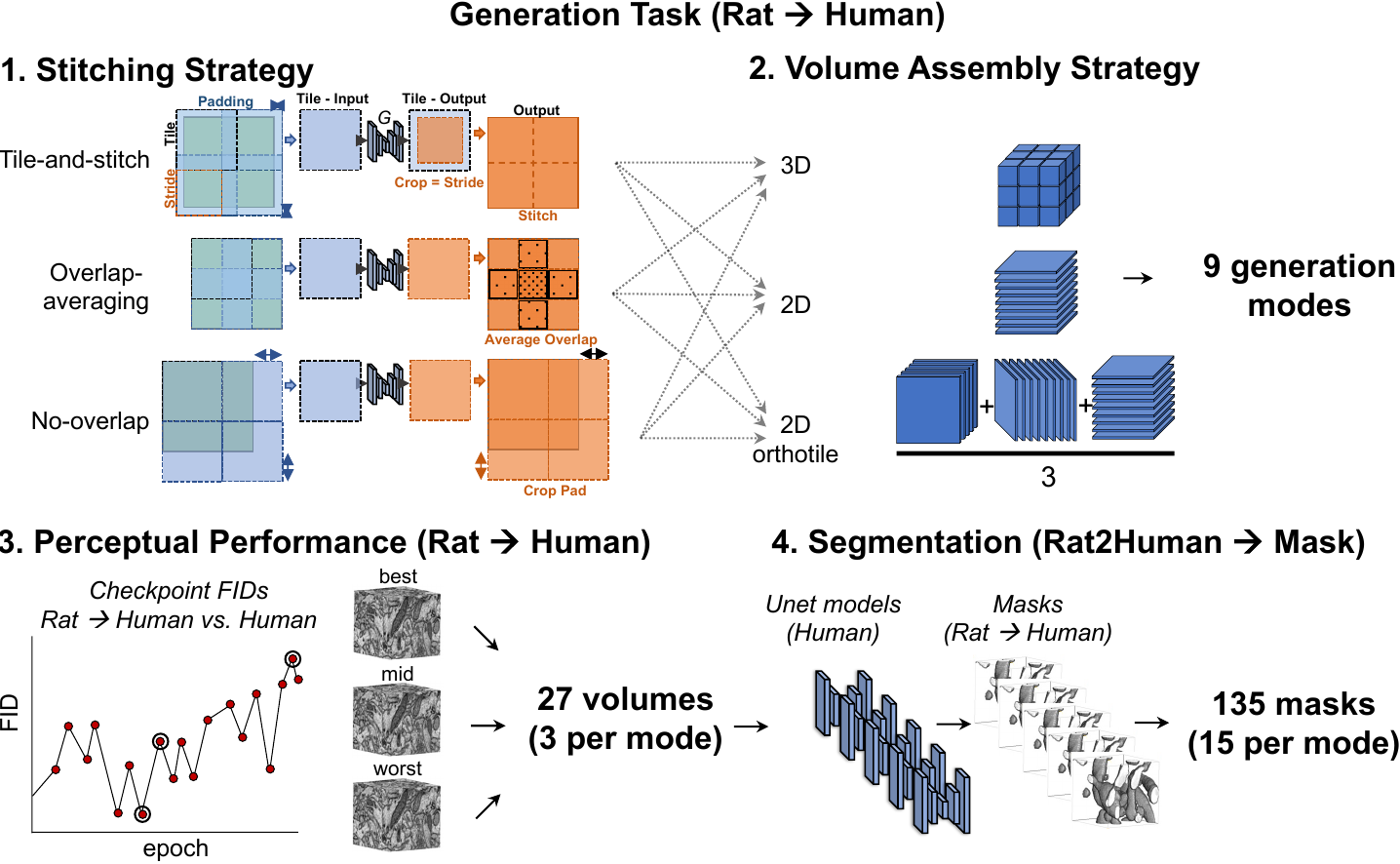}
    \caption{\textit{Experimental Design: 1. Three different stitching methods are compared. 2. Three different types of volume assembly are used, combined with the stitching methods. 3. For each mode of assembly (stitching x volume assembly), volumes are generated for each training checkpoint of the generator (rat-to-human) model and FID wrt ground-truth (human) domain is determined. Out of 20 checkpoints, the three highest, lowest and three mid-range by FID are chosen for downstream analysis. 4. Five Unet models are trained with different losses and metrics on the mitochondrial segmentation task of the human EM volume. Each model is tested for the segmentation of the volumes chosen from the model checkpoints from the rat-to-human volume assembly.}
    }
\label{fig:experiment}
\end{figure*}

\begin{figure*}[h]
\centering
\captionsetup{width=0.89\textwidth}
\includegraphics[width=0.89\textwidth]{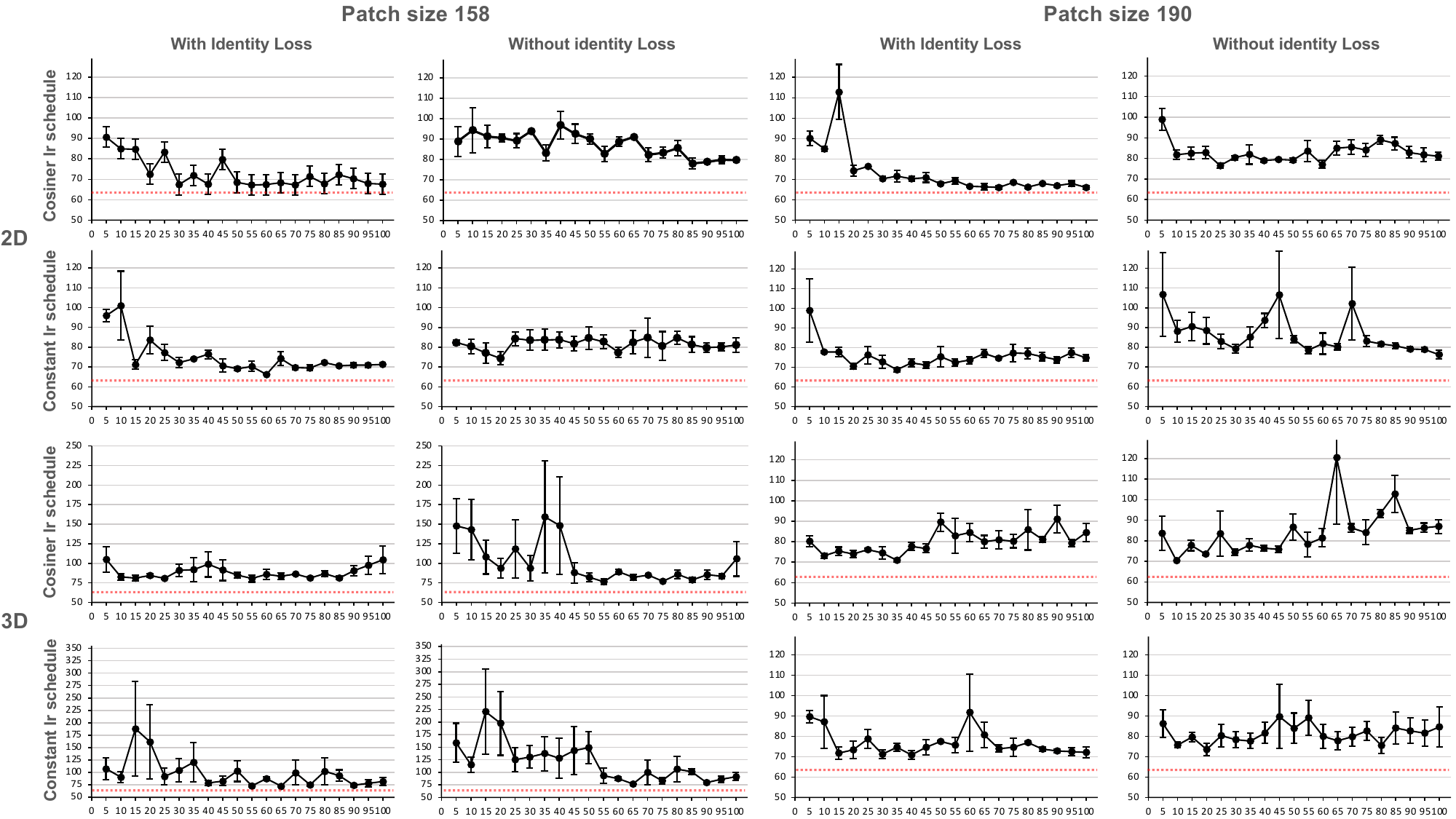}
\caption{\textit{Effect of parameter choice on FID: Shown are the FID scores for each generated checkpoint during training of the openorganelle dataset, in the direction HeLa-to-jurkat, when compared to the GT openorganelle jurkat dataset. Marks represent the average FID from a triplicate of each checkpoint, error bars indicate one standard deviation around these means. The dotted line represents the baseline FID (63.93) between the ground-truth datasets (HeLa and jurkat)}}
\label{fig:openorganelle}
\end{figure*}

The main experiment to measure perceptual and downstream effects of the stitching and dimensionality choices was performed using the \textit{MitoEM} dataset (see \ref{subsubsec:mitoem}). Prior to this experiment, we carried out control experiments on a dataset from the \textit{openorganelle} database (see \ref{subsubsec:openorganelle}) which was designed to aid in a suitable choice of training hyperparameters. Specifically, we tested different input patch sizes and dimensions, the use of identity loss and different learning rate schedules while measuring perceptual performance for whole training runs in cycleGAN (Supplementary Fig. \ref{fig:openorganelle}). The predicted volumes in this preliminary experiment were generated using the tile-and-stitch approach \cite{rumberger2021shift}.
Based on these results, the cycleGAN models were trained with identity loss, a constant learning rate and with the largest patch size fitting into GPU-RAM, with a side-length of 190 pixels/voxels.
Using the \textit{MitoEM} dataset, we trained a cycleGAN model to translate between rat and human EM stacks, using the full datasets. The trained model was then used to generate rat-to-human translated EM volumes, using the three above-mentioned stitching strategies (see Fig. \ref{fig:experiment}, 1.):
\begin{enumerate}
    \item tile-and-stitch. In this approach generator models with valid convolutions (with no padding) are used in training and inference. Output patches are cropped according to the tile-and-stitch principle \cite{rumberger2021shift}. Given an input patch size of 190x190 pixels or 190x190x190 voxels and 5 downsampling steps with a convolution of stride 2, gives an output-crop size of 64 pixel/voxel edge length for each output patch. During inference, neighbouring input patches are shifted by exactly the same amount. These cropped output patches are then assembled without overlap. Padding is added to the input volume, such that cropped output patches along the image edges align with the input volume. Excessive padding after stitching the full volume is cropped to ensure that the final volume has the same size as the input volume.
    \item Padded convolutions with averaged overlap. Here, the same model was used, but the convolutions were padded with a row of zeros during inference. A "Halo" was added to the patches, with reflective padding, with 16 pixels on each side of the patch, which we found improved overall quality of the predictions. After inference of each patch the halo pixels were removed. This effectively made the input patch side length 222. After inference, the halo was removed, giving a patch size of side-length 190, and the patches assembled with an overlap of 32 pixels. The assembled image was then divided by a normalisation matrix which adjusted the pixel intensity according to the number of patches which contributed to each pixel coordinate on the image.
    \item Valid convolutions were used during inference as in (1), but the output patches were neither cropped nor was the step size between neighbouring windows adjusted. Instead the step size was kept the same as the output patch size, effectively removing any overlap between patches. The input and output patch side length was 190 pixels.
\end{enumerate}

The stitching methods were tested with 2D and 3D patches and respective models of matching dimensionality. In the 2D models, the volumes were assembled in the z-direction slice-by-slice. Additionally, we tested an 'orthoslice' approach which effectively creates the volume as an ensemble of 2D inferences in three sectional directions (Fig. \ref{fig:experiment},  2.). We compared the outputs of the rat-to-human model visually and via FID to the human dataset (see \ref{subsec:fid-calculation}).
For the evaluation of downstream segmentation, we used the best, worst and a mid-range checkpoint, ranked by their respective FID (Fig. \ref{fig:experiment}, 3.).

Downstream performance based on different assembly strategies was evaluated by using five Unet models trained with different losses and evaluation metrics on the in-domain segmentation of human mitochondria. Each model was then tested on the rat-to-human generated volumes from the above GAN models (Fig. \ref{fig:experiment}, 4.). The thresholded probability maps were scored via full-image Intersection-over-Union (IoU) with respect to the ground-truth mask in the rat-domain. 

\section{Results}

\subsection{Perceptual Effects}

In the initial experiments on the \textit{openorganelle} dataset, we found that all models converged relatively slowly according according to FID. Indeed, under most parameter settings the models did not improve the perceptual distance between the jurkat and HeLa datasets (FID: 63.93). Nonetheless, there were clear differences in the stability of training. Overall, 2D models were more stable both during individual training runs and between triplicates with different initialisations. We also found that the identity loss improved stability, which was specifically notable when the models were trained on larger patch sizes and in 3D. The exact patch size had little impact on the average performance of the models, though it reduced the variance between different model initialisations, thus improving reproducibility between runs.

From the initial evaluation between the human and rat dataset, we find the FID between the datasets to be  56.60. The FID between the best predicted volumes and the targets from all stitching and dimensionality conditions range between 30 and 40 (table \ref{tab:perceptual-effects}  and supplementary tables 2-10), suggesting that style transfer in this task can lead to an overall improvement in the similarity between datasets.
Evaluating FID for predictions using different patch dimensionalities, we firstly find that 2D models perform better than 3D models (table \ref{tab:perceptual-effects}). 

\begin{table}[ht]
\centering
\begin{tabularx}{0.88\textwidth}{
  | >{\raggedright\arraybackslash}p{1.2cm}
  || >{\centering\arraybackslash}X
  | >{\centering\arraybackslash}X
  | >{\centering\arraybackslash}X
  | >{\centering\arraybackslash}X
  | >{\centering\arraybackslash}X
  | >{\centering\arraybackslash}X
  | >{\centering\arraybackslash}X
  | >{\centering\arraybackslash}X
  | >{\centering\arraybackslash}X| }

    \hline
    \textbf{Model} &
      \multicolumn{3}{c|}{\textbf{3D}} &
      \multicolumn{3}{c|}{\textbf{2D}} &
      \multicolumn{3}{c|}{\textbf{2D (orthoslice)}} \\
    \hline
    \textbf{\textit{Stitching}} &\textit{T\&S} &\textit{Valid} & \textit{Same} &\textit{T\&S} &\textit{Valid} &\textit{Same} &\textit{T\&S} &\textit{Valid} &\textit{Same}\\
    \hline
    \textbf{FID $\downarrow$} & 38.99 & 36.99 & 41.21 & 30.49 & \textbf{30.23} & 30.45 & 40.74 & 40.71 & 41.02\\
    \hline
    \textbf{IoU $\uparrow$} & 0.7135 & 0.7155 & 0.7120 & 0.7218 & 0.7230 & 0.7220 & 0.6910 & 0.7304 & 0.7308  \\
     \hline
\end{tabularx}
\captionsetup{width=0.89\textwidth}
\caption{Assembly modality vs. Perceptual Performance. Only the results for the models yielding the best perceptual performance (by FID) with respect to the target domain (human) are shown. The IoU values are given for reference. For full results, see Supplementary Tables (\ref{sec:supplementary-tables}}
\label{tab:perceptual-effects}
\end{table}

Across the full range of FID scores during training (from best to worst), scores tend to be better in 2D models than in 3D models. The results from the orthoslice model appear to be in between 2D and 3D models in terms of their perceptual scores by FID. These results are not unique to this task, but were already reflected in the initial hyperparameter tuning experiment (see supplementary results), where 3D models tend to show more noisy performance throughout training, giving less consistent FID scores than 2D models. Another observation from this experiment was the positive effect of identity loss to the training which stabilized performance during training and for different weight initialisations. Of the two types of learning rate schedules tested in the control experiment neither had a strong positive effect on training.
Stitching had a small effect on FID, with the perceptual performance tending to be better for volumes generated by valid-padded convolutions in the backbone and without tile overlap. Overlapping tiles during stitching and using same-padded convolutions did not affect the overall performance significantly for the 2D model but seems to have a negative effect on the 3D model. The scores for the models used for orthotile generation were extremely close in all conditions, apparently overshadowing most differences resulting from different stitching methods.

\begin{figure*}[!ht]
\centering
%\begin{figure*}[h]
    \captionsetup{width=0.89\textwidth}
    \includegraphics[width=0.89\textwidth]{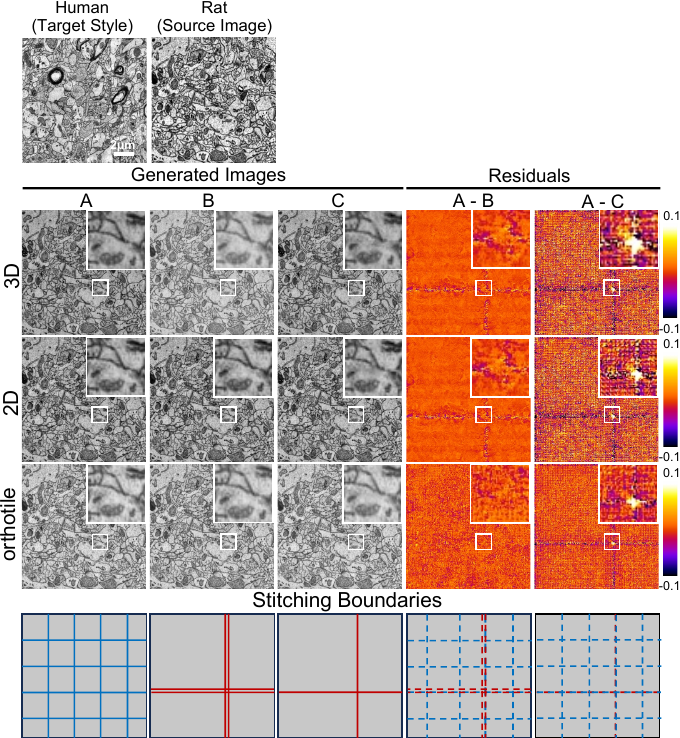}
    \caption{\textit{Stitching approaches and generated images - Top: Example target (Human FIB-SEM) and input (Rat FIB-SEM). Middle (Left): "Generated Images" with different stitching methods: A) Tile-and-Stitch B) padded convolutions with overlap C) Valid convolutions, no overlap. "Residuals": These are constructed by subtraction of the normalised images from each other, with A-B and A-C, referring to the images in columns A to C. Bottom: An illustration of the patch boundaries in the images in each column.}}

\label{fig:perceptual-effects}
\end{figure*}

We aimed to visualise potentially introduced artefacts in the generated images. To this end, we created residuals between images generated via tile-and-stitch (known to be artefact-free)(Fig. \ref{fig:perceptual-effects}, A), overlapping and averaged patches (Fig. \ref{fig:perceptual-effects}, Residuals: A-B) or patches without overlap from a model with valid-padded convolutions (Fig. \ref{fig:perceptual-effects}, Residuals: A-C). All residuals were created by normalising the example crops to a range between 0.0 and 1.0 in float32 precision and calculating the absolute difference between the images.
The examples in figure Fig. \ref{fig:perceptual-effects} show 300x300 crops in the xy plane of the generated volumes. By eye, stitching artefacts are virtually unnoticeable in all cases. However, the residuals reveal that stitching artefacts are present in the overlap and valid-padding conditions, given that clear lines are visible at the boundary of the patches used to generate the volumes for these conditions (190 pixels), but not at the boundary of the cropped patch used in the tile-and-stitch approach (64 pixels). The difference as estimated by the residuals is clearer in the condition without overlap (Fig. \ref{fig:perceptual-effects}, A-C) than with overlapping and averaged patches (Fig. \ref{fig:perceptual-effects}, A-B). The differences between the different dimensionalities are visually minor, with the only exception being the residual constructed from the volumes generated using orthotiles. Here, the stitching artefacts are extremely faint when the volumes were generated with overlapping patches and padded convolutions.

\newpage

\subsection{Effects on downstream segmentation}
Visually, the impact of stitching and model/patch dimensionality is minor. Do these subtle differences impact the performance of downstream models? To determine which models' predictions produced the best results in the segmentation, we compared cycleGAN checkpoints which yielded different perceptual scores (FID) during training. The best segmentation results  for each assembly mode are shown in table \ref{tab:segmentation-effects}. The full results of the experiments are shown in the supplementary material (\ref{sec:supplementary-tables})

Independent of the stitching approach, 3D assembly tends to peak at slightly higher segmentation scores than the 2D and orthotile models, particularly when using the binary cross-entropy loss to train the Unet models. When other losses were used for training, the advantages of the 3D model disappear almost entirely. When comparing only the stitching approaches, tile-and-stitch assembly peaks at slightly higher values in the downstream segmentation of these volumes than for volumes assembled using the other stitching approaches. This effect is more robust across different training settings of the segmentation model than the dimensionality. The other two types of stitching do not differ notably with respect to downstream segmentation performance. Segmentation on orthotile-assembled volumes is often better than on 2D generated volumes. The only outlier is the orthotile assembly in tile-and-stitch mode, which appears to deteriorate. However, we could not rule out issues in the pipeline causing this result. 

Following these trends for downstream segmentation performance, the 3D model combined with the tile-and-stitch approach led to the best performance for downstream segmentation (Table \ref{tab:segmentation-effects}, 3D T\&S). However, the overall differences between downstream segmentation results from different dimensionalities and stitching methods are very small, often differing only in the 3rd or 4th decimal point.

A notable insight from the full results (Supplementary \ref{sec:supplementary-tables}), which the results from Table \ref{tab:segmentation-effects} are derived from, is that the perceptual metric (FID) is not sensitive to the differences which yield better downstream performances for the segmentation tasks. The volumes with the best FID with respect to the target domain almost never result in the best segmentation scores, although the worst FID scores often correlate with lower downstream segmentation performances.

\begin{table}[ht]
\centering
\begin{tabularx}{0.88\textwidth}{
  | >{\raggedright\arraybackslash}p{1.2cm}
  || >{\centering\arraybackslash}X
  | >{\centering\arraybackslash}X
  | >{\centering\arraybackslash}X
  | >{\centering\arraybackslash}X
  | >{\centering\arraybackslash}X
  | >{\centering\arraybackslash}X
  | >{\centering\arraybackslash}X
  | >{\centering\arraybackslash}X
  | >{\centering\arraybackslash}X| }

    \hline
    \textbf{Model} &
      \multicolumn{3}{c|}{\textbf{3D}} &
      \multicolumn{3}{c|}{\textbf{2D}} &
      \multicolumn{3}{c|}{\textbf{2D (orthoslice)}} \\
    \hline
    \textbf{\textit{Stitching}} &\textit{T\&S} &\textit{Valid} & \textit{Same} &\textit{T\&S} &\textit{Valid} &\textit{Same} &\textit{T\&S} &\textit{Valid} &\textit{Same}\\
    \hline
    \textbf{IoU $\uparrow$} & \textbf{0.7383} & 0.7359 & \textbf{0.7382} & 0.7374 & 0.7230 & 0.7372 & 0.6927 & 0.7334 & 0.7325  \\
     \hline
    \textbf{FID $\downarrow$} & 43.52 & 40.91 & 47.85 & 37.24 & 30.23 & 37.07 & 42.53 & 42.53 & 42.50\\
    \hline

\end{tabularx}
\captionsetup{width=0.89\textwidth}
\caption{\textit{Assembly modality vs. Segmentation Performance. Only the results for the models which produce the best segmentation masks (by IoU) with respect to the ground-truth segmentation masks are shown. The FID values are given for reference. For full results, see Supplementary Tables. }}
\label{tab:segmentation-effects}
\end{table}

\begin{figure*}[!ht]
\centering
%\begin{figure*}[h]
    \captionsetup{width=0.89\textwidth}
    \includegraphics[width=0.89\textwidth]{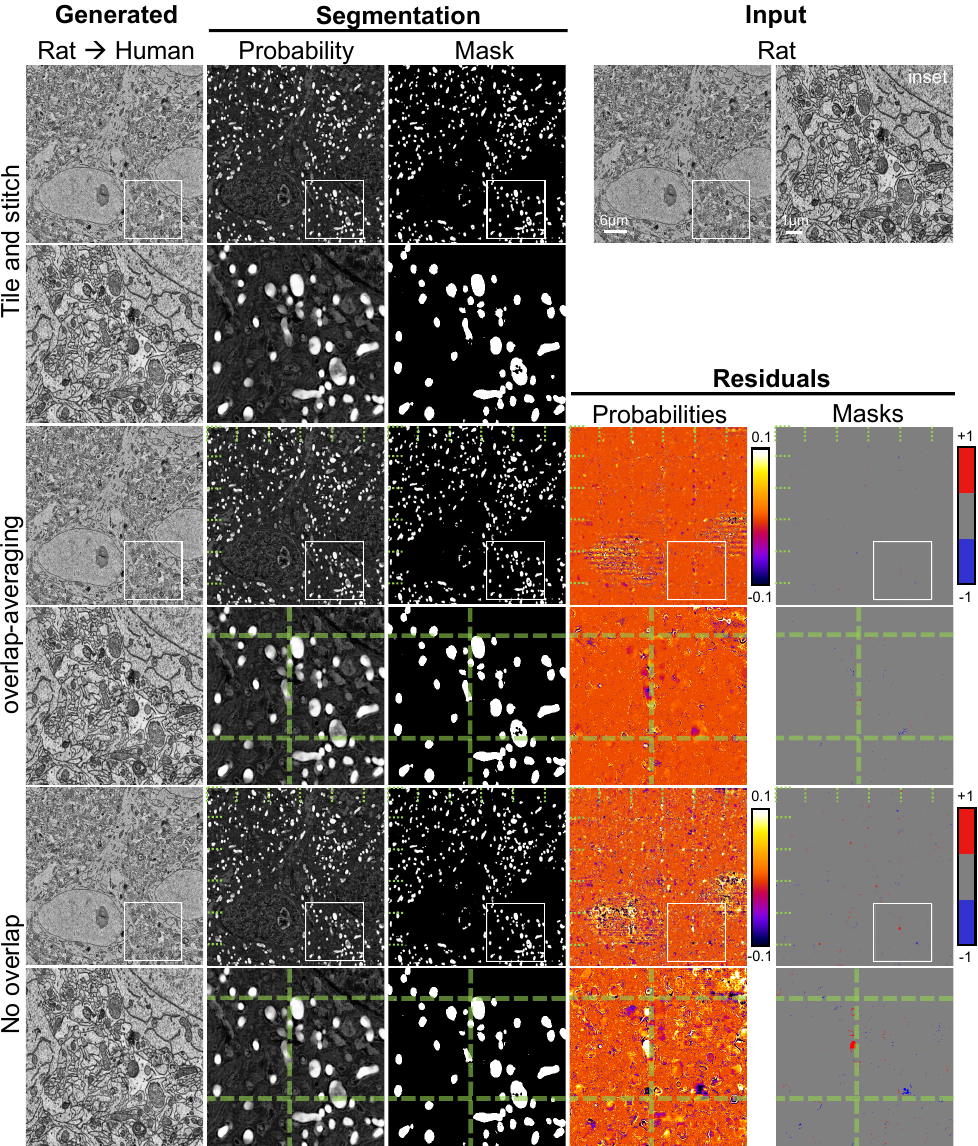}
    \caption{\textit{Effect of stitching on downstream segmentation - Probability maps, masks generated by thresholding and overlays with the input image are shown, in rows 1, 3 and 5 for a full slice of the input volume and in rows 2, 4 and 6 for the insets. The green lines indicate the edges of the tiles used to assemble the input volume. The residuals represent the differences between the respective stitching type (shown on the left) and the tile-and-stitch}}
\label{fig:segmentation-effects}
\end{figure*}

We next visualised segmentation masks for the different stitching methods. Figure \ref{fig:segmentation-effects} shows probability maps and segmentation masks for the example of volumes generated by a 2D cycleGAN model with slice-by-slice assembly. We first tested whether stitching artefacts were still present after the generated volumes had been processed by the unet (Fig. \ref{fig:segmentation-effects}, "Probability"). Though not visible in the probability maps themselves, when comparing predictions of overlapped-averaged stitching and no-overlap stitching with the artefact-free volumes generated via tile-and-stitch (Fig. \ref{fig:segmentation-effects}, "Residuals - Probabilities"), the stitching artefacts are visible along the patch sizes used in both the former stitching approaches.
The stitching artefacts are not visible in the predicted masks by eye (Fig. \ref{fig:segmentation-effects}, "Mask"). When generating residuals between masks generated on artefact-free stitched and the other volumes (Fig. \ref{fig:segmentation-effects}, "Residuals - Masks"), the differences are minor. However, the differences between the masks appear to be larger and more frequent in proximity the stitching boundaries, and align with the strongest differences in the residuals of the probability maps. These differences are more visible in the residuals using images generated without patch overlap than in those with overlapping patches.

\begin{figure*}[!ht]
\centering
    \captionsetup{width=0.89\textwidth}
    \includegraphics[width=0.89\textwidth]{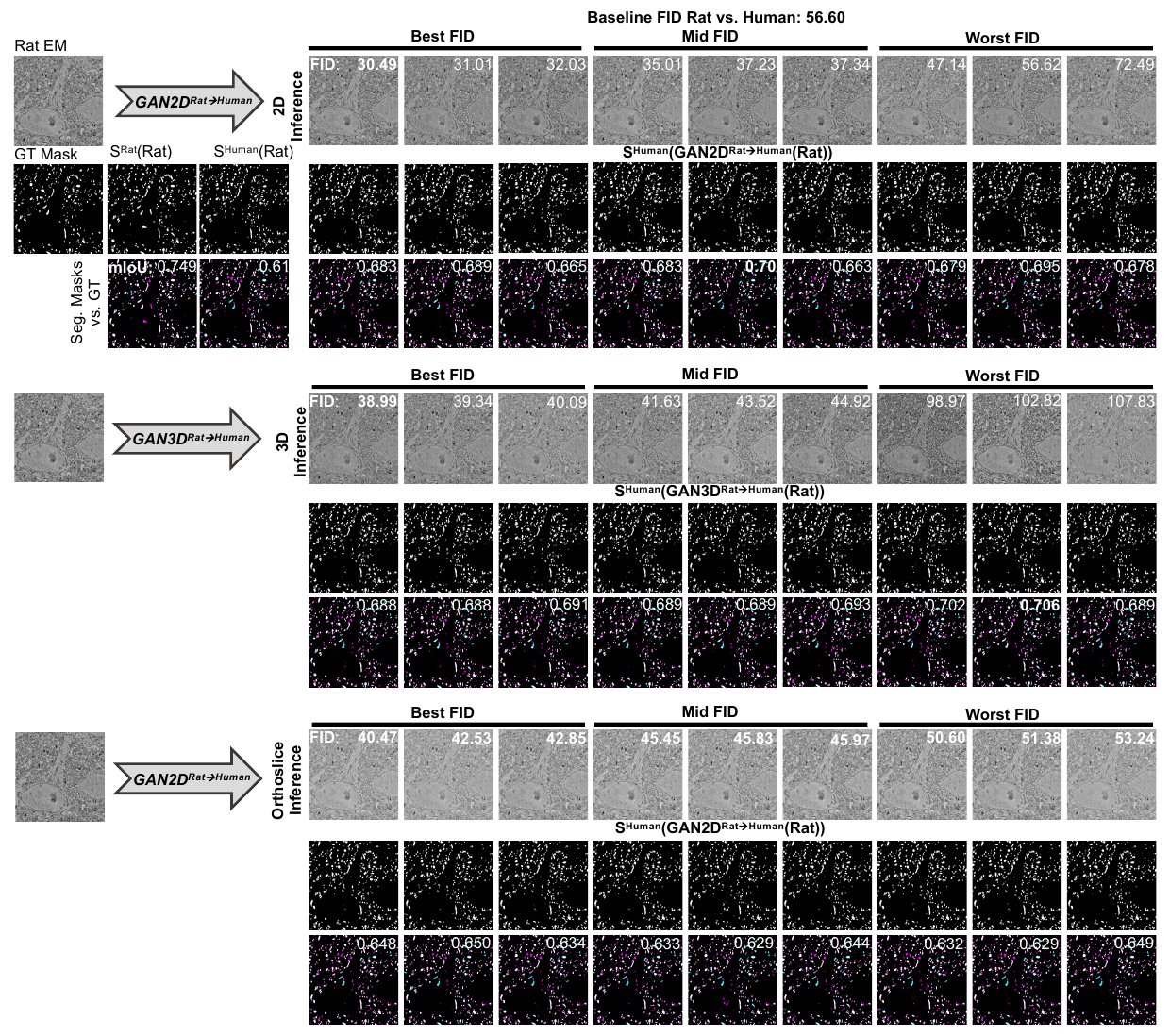}
        \caption{\textit{Comparison of segmentations between different dimensionality settings, in tile-and-stitch: A rat FIB-SEM volume is transformed into the human FIB-SEM domain via three inference settings with different respective assembly strategies (2D, 3D or orthoslice). 9 checkpoints (3 best, 3 mid-range, 3 worst) were chosen for downstream segmentation performance analysis. FID scores are shown on the generated images, and IoU scores on the segmentation masks. (Legend for segmentation masks: white=overlap with GT, magenta=false positive, blue=false negative, Symbols: S\textsuperscript{Rat}(Rat) refers to a segmentation model trained on rat FIB-SEM volumes and used for inference on rat FIB-SEM, S\textsuperscript{Human}(Rat) refers to the segmentation model trained OOD on human FIB-SEM volumes and used for inference on the rat-FIB-SEM dataset.}}
\label{fig:dimensionality}
\end{figure*}

We also compared the differences in assembly dimensionality visually. (Supplementary Figure \ref{fig:dimensionality}) Out of the selected sample of models the best performance by IoU on mitochondria was achieved by a 3D model. Interestingly, this relatively high score was achieved on a volume that had one of the worst FID scores with respect to the target volume. The visualisations did not give insights into the differences between the segmentations, i.e. where the best 3D models performed superior to the best 2D models.

%\end{figure*}

\clearpage
\section{Discussion}

\subsection{Perceptual effects}

Perceptually, according to the FID metric, 2D models' outputs were closer to the target domain than 3D models'. These results were robust, regardless of the stitching approach. A likely explanation for this positive effect is the larger batch size used during training of the 2D models, but not the 3D models where the batch size is limited to 1 during training. It appears the advantage of additional context in 3D slices does not outweigh the advantage of a larger variance within a batch to learn the style-transfer. The batch-effect may also explain the presence of extreme outliers in the FID scores for the 3D models, where the batch size of one makes the training inherently more noisy and leads to a larger variance between checkpoints.

Orthotile assembly usually increased (worsened) the best FID scores for the 2D models, but improved the perceptual quality of the worst checkpoints, thus 'smoothing' extreme points in FID scores. This is likely a result of ensembling the predictions, which may reduce the effect of artefacts present in only one sectional view, but which conversely could also diminish the effect of style features in the best 2D models, by averaging the best predictions with slightly worse ones in a different sectional direction.

The effect of stitching on perceptual performance is not clear. Considering the best checkpoints for the 2d and 3d models, it is clear that the presence of stitching artefacts in the data does not diminish perceptual quality according to this metric. Instead, for 2d models, there seems to be a neglible negative effect on perceptual performance. Thus, even when artefacts are present, as we show through the residual maps of the predictions, the FID does not account for this when evaluating image quality. This means that during the generation of large images via tiling, metrics beyond perceptual ones such as FID should be considered when determining the generation quality. This is important since we showed in the other experiments that these artefacts can indeed propagate to downstream tasks.

\subsection{Effects on downstream segmentation}

Though the performance of 3D models for the style transfer task is slightly inferior to that of 2D models, the overall best performance in an OOD segmentation task of mitochondria was achieved by a 3D model. 
Interestingly, 3D models appear to be superior only when used with artefact-free stitching. When other types of stitching are used, 2D slice-by-slice and 2D orthotile assembly have a similar or even better performance than when these stitching methods are used with 3D models and patches. This might suggest that overlapping and averaged boundaries of patches, or mismatching borders have a stronger negative effect on downstream tasks when the assembling patch of the generated image is a 3D patch. At present, it is not clear what causes this effect, and whether this difference even arises from differences in stitching alone.

Though the residuals we used to visualise the discrepancies between different generated probability maps and masks show that the differences in segmentation performance are indeed likely to be related to the stitching boundaries in the GAN-generated images, they also show that other areas of the images can contribute to these differences. One source of these other discrepancies could lie in the patch sizes used during inference. For instance, the edge-length of the patches used for the tile-and-stitch models was 192, whereas for the model with same-padded convolutions and patch overlap it was 190, which may result in small discrepancies between the patches which are unrelated to the stitching approach. Furthermore, as the UNet is not shift equivariant, any patch offset that is not a multiple of the periodicity of the shift equivariance can result in different output -- consequently, discrepancies can emerge even with the identical input- and output patch size if tiling is performed at an offset.

The orthotile assembly method for the generated images represented an experimental approach. Could a relatively naive ensemble of generated predictions lead to an improvement of the generated image? Interestingly, the answer to this question appears to be two-fold. When the tile-and-stitch approach was used, the orthotile-assembly method led to worse downstream segmentation results (compare Supplementary Tables  \ref{tab:Tile-and-Stitch - 2D} and \ref{tab:Tile-and-Stitch - 2D (orthoslice)}), while for the other conditions, it achieved similar (compare Supplementary Tables \ref{tab:Same-padding - 2D} and \ref{tab:Same-padding - 2D (orthoslice)}) or even stronger results (compare Supplementary Tables \ref{tab:Valid-padding, no crop - 2D} and \ref{tab:Valid-padding (no crop) - 2D (orthoslice)}). 

Though it is not clear why the tile-and-stitch approach leads to an inferior result when images are assembled in this way, in the latter case, this could be a result of averaging the stitching boundaries which tend to be slightly more prevalent when images are generated without overlapping patches (see Fig. \ref{fig:perceptual-effects}). Since the volumes are assembled from three directions, most pixels/voxels lying on stitching boundaries will be the average of only one underlying 'stitching' pixel/voxel and two pixels/voxels which are not on a stitching boundary. This is likely to diminish the effect of stitching, in conditions where no overlap is used between neighbouring tiles. However, the effect appears not to be compounded when neighbouring patches are already averaged via overlap.

\subsection{Limitations}

We recognize some limitations in this study, which includes the choice of the downstream task, a relatively simple foreground-background segmentation, which made it difficult to analyse the small differences resulting from the image generation approaches used in this study. The tile-and-stitch method is also currently limited to relatively simple convolutional Unet-style architectures. Future research should consider how stitching can be mitigated in more sophisticated architectures, such as those using attention blocks, and on other downstream tasks, such as instance segmentation. Furthermore, comparing the perceptual similarity between the images has significant shortcomings. FID, as many of the other perceptual image metrics for unpaired image datasets, is based on image features from natural images. It is unclear how well this metric can capture the differences between images from the EM domain at all. Although our results show somewhat intuitively that GANs improve the visual similarities between source and target domain images according to FID, the mechanism for FID identifying these similarities or differences remains opaque. A more insightful metric for image-features in tasks on biological datasets would be one that is based on features from such datasets, i.e. EM datasets. This would make the results from perceptual comparisons via image metrics more easily interpretable.
Finally, unlike for the segmentation masks, we did not investigate whether the differences between the perceptual scores for different stitching methods were a result of the presence of stitching scars or not. FID appears to be able to identify if images contain significant artefacts from the GAN process, which is the likely source of some of the outliers seen in the experiments, but it is not clear how it distinguishes between images which have relatively few distinct GAN-related artefacts.

\section{Conclusions}

In this study, we set out to identify the impact of previously understudied aspects of GAN performance for style-transfer of large EM volume datasets, namely, the effects of stitching and dimensionality on perceptual quality and the performance of downstream segmentation.

The stitching and dimensional approaches tested in this study differ only slightly in their effect on perceptual and downstream segmentation metrics and visually. Nonetheless, our key finding is that these differences are big enough to show that 3D models, applied with an artefact-free stitching approach, can outperform 2D models in generating images which optimise performance of downstream OOD semantic segmentation. Through comparisons between generated images, we also show that these improvements are indeed most likely a result of the reduction in stitching scars. The improvements seen in 3D models are marginal, and given the additional computational cost of training and storing the weights of 3D models, are likely not worth the computational cost. However, if performance must be optimised, an artefact-free stitching approach and the use of a 3D model, is likely to yield performance benefits.

We also make several additional observations which could guide users in training models and assembling large 3D images using style transfer:

\begin{itemize}
    \item The choice of checkpoint can significantly impact the perceptual and downstream performance of style-transfer models. Here, 2D models offer the benefit of being generally more stable during training, owing to relatively low cost of training with larger batch sizes. Thus, individual checkpoints of 2D models are more likely to yield near-optimal performance. 
    As computational resources improve, it could become more feasible to train 3D models with larger batch sizes, which might stabilize and improve the training of these large models, and increase the performance benefits of using such models.
    \item Our study is, to our knowledge, the first to investigate how ensembling of generated images impacts downstream metrics, and we find that it can benefit the performance of a downstream OOD task, especially when the underlying checkpoints initially contain GAN-related artefacts or have a low perceptual performance. However, the additional cost of ensembling an image from different directions needs to be considered. In this study, we also used datasets which were (made) isotropic in all dimensions to generate orthotile ensembles, which may limit this method to datasets with very high axial resolution, such as the cryo-EM tomograms used here.
    \item As was proposed in previous studies, but which we empirically showed in this research, the relation between the perceptual performance of generative models, and their effect on downstream performance for another task (such as semantic segmentation) is weak. For domain-adaptation tasks in biological images, this means that generative models must either be evaluated on downstream tasks, or more insightful perceptual metrics must be developed for tasks in which style-transfer is used for unpaired biological data.

\end{itemize}

Our study represents a first attempt to investigate the impact of image assembly in the overall quality of generated images in bioimage analysis. The majority of previous research in image generation tasks has investigated the quality of model predictions on individual patches with relatively limited dimensions. Our study highlights that in domains where much larger images are common, such as bioimaging, other aspects such as image dimensionality and the stitching approach for the full image, should be taken into account when assessing the quality of a generative model.

An additional outcome of this work is a comprehensive reworking of the original cycleGAN codebase, to allow a transparent stitching approach for inference of large EM volumes. This includes an improved pipeline for data-loading, including in zarr-format, which allows training and inferring on data that is virtually unlimited by data-size. The codebase is designed to facilitate and improve the community's expertise with stitching techniques, but fundamentally provide a tool for style-transfer for very large datasets for the bioimaging community.

\section*{Acknowledgment}
This work was funded by Helmholtz Imaging Project ZT-I-PF-4-025 \emph{SyNaToSe}.

\newpage

\section{Methods}

\label{sec:methods}

\subsection{FID score calculation}
FID scores were evaluated using the pytorch-fid repository \cite{Seitzer2020FID}. The rat-to-human models were evaluated on the training volumes directly. For each training checkpoint, a prediction was created in the human target domain. The FID was then computed by dividing the predicted volume and the training volume in the respective target domain into 20000 patches of shape 128x128 pixels. This patch size ensured that, given a cycleGAN patch size of 190 or 192, that any patch boundaries in the generated images would usually lie within the patches sampled for FID, ensuring that they could affect the score. The FID was calculated with a batch size of 50, with 2048 feature dimensions.
\label{subsec:fid-calculation}

\subsection{Procedure to achieve seamless stitching in Unet Generators}
Artifact-free stitching was achieved as follows:
The use of valid convolutions in a convolutional layer means that the size of the feature map dimensions are reduced by two pixels/voxels in each downsampling step. Using the original cycleGAN's Unet, which uses convolutions with stride 2 and no pooling, means that the dimensions of the feature maps after each downsampling step are:

\begin{equation}
    d^{down}_o = (d^{down}_i - 2) / 2
\end{equation}

with $d^{down}_o$ and $d^{down}_i$ the sizes of the output and input dimensions of the feature maps in the downsampling layer respectively. Given an input patch size of 190, and 5 downsampling steps, this gives a bottleneck size of the feature maps of 4. When the patches are upsampled, the outer pixels/voxels now must be discarded in each upsampling block, to ensure that only pixels/voxels present in the upsampling layer input contribute to the outputs. For the upsampling layers, the following therefore holds:

\begin{equation}
    d^{up}_o = (d^{up}_i \times 2) - 2
\end{equation}

Starting at the bottleneck of $d^{up}_i = 4$, after 5 upsampling blocks, the feature map's cropped size is 66. In order to ensure artefact-free stitching as proposed in Rumberger et al. \cite{rumberger2021shift}, the output should be further cropped to a multiple of $f^N$, where $f$ is the downsampling factor and $N$ the number of downsampling steps. For the given model, this means a multiple of $2^5 = 32$. The output tiles are therefore cropped to the nearest multiple for the given output crop, i.e. to size 64. Note that without cropping, the upsampled images of the cycleGAN's Unet generator have the same size as the input. This 'no-crop' setting was used for the 'no-overlap' condition in the experiments above.

To stitch an inferred image artefact-free, the stride between neighbouring patches should match exactly the cropped patch size calculated above. Since the resulting image would be smaller than the input, due to the cropped patches, we pad the input stack (via mirror padding, although other padding options can also be used) to ensure that the output stack will have the same size as the input. The padding at 0 ($p_init$) in all dimensions is chosen to be the same, such that the central crop of the first patch, in this case of size 64, according to the above considerations, overlaps exactly with the corner of the original (unpadded) image. We now determine the number of patches needed to cover the full size of the given thus-padded dimension of the stack, with the stride between patches given by the output crop size (see above). The number of patches along the dimension is thus determined as:

\begin{equation}
    N_i = ((V_i + p_{init} - P_i) / S_i ) + 1
\end{equation}

where $V_i$ is the size of the volume along dimension $i$ and $P_i$ and $S_i$ are the input patch size and the stride along the same dimension, respectively.

Given the number of patches and the stride, the new volume size needed to ensure patching without overhanging pixels/voxels is:

\begin{equation}
    V^{new}_i = (N_i \times S_i) + P_i
\end{equation}

and the required padding along the dimension $i$ can be easily calculated as the difference between the original volume (with initial padding) and the new volume.

The padded volume can now be divided into patches of the input patch size with the stride equal to the output crop, while ensuring that the entirety of the input volume contributes to the output. The output volume can now be assembled from the cropped output patches processed by the Unet generator, without overlap between patches.

\newpage

\section{Supplementary Results}

\subsection{Supplementary Tables}

The supplementary tables show segmentation results for additional checkpoints from the cycleGAN generators. Shown are the best, worst and a mid-range checkpoint by FID (final column). The values in each column are the mean IoU scores with respect to the ground-truth segmentation after thresholding the probability maps. The headings in each column represent the loss and the validation metric used during training of the Unet segmentation model. 
\label{sec:supplementary-tables}

\begin{table*}[ht]
\centering
\setlength{\tabcolsep}{2pt}
\begin{tabular}{
  | >{\raggedright\arraybackslash}p{4cm}
  | R{1.45cm} 
  | R{1.45cm} 
  | R{1.44cm} 
  | R{1.58cm} 
  | R{1.58cm}
  | R{1cm} |
}

\hline
\textbf{Unet-models: \textit{Loss, Metric}} 
& \multicolumn{1}{c|}{\textit{Dice, IoU}} 
& \multicolumn{1}{c|}{\textit{MSE, IoU}} 
& \multicolumn{1}{c|}{\textit{BCE, IoU}} 
& \multicolumn{1}{c|}{\textit{Dice, MSE}} 
& \multicolumn{1}{c|}{\textit{MSE, Dice}} 
& \multicolumn{1}{c|}{\textbf{FID}}\\
\hline

\textbf{Best FID checkpoint}  & 0.6876 & 0.6920 & 0.7135 & 0.6827 & 0.6281 & 38.99\\
\hline
\textbf{Mid FID checkpoint}  & 0.7024 & 0.7199 & \textbf{0.7383} & 0.7156 & 0.6203 & 43.52\\
\hline
\textbf{Worst FID checkpoint}  & 0.6894 & 0.6284 & 0.7259 & 0.7009 & 0.6164 & 107.83\\
\hline
\end{tabular}
\caption{Tile-and-Stitch - 3D}
\label{tab:Tile-and-Stitch - 3D}
\end{table*}

\begin{table*}[ht]
\centering
\setlength{\tabcolsep}{2pt}
\begin{tabular}{
  | >{\raggedright\arraybackslash}p{4cm}
  | R{1.45cm} 
  | R{1.45cm} 
  | R{1.44cm} 
  | R{1.58cm} 
  | R{1.58cm}
  | R{1cm} |
}

\hline
\textbf{Unet-models: \textit{Loss, Metric}} 
& \multicolumn{1}{c|}{\textit{Dice, IoU}} 
& \multicolumn{1}{c|}{\textit{MSE, IoU}} 
& \multicolumn{1}{c|}{\textit{BCE, IoU}} 
& \multicolumn{1}{c|}{\textit{Dice, MSE}} 
& \multicolumn{1}{c|}{\textit{MSE, Dice}} 
& \multicolumn{1}{c|}{\textbf{FID}}\\
\hline
    \textbf{Best checkpoint by FID}  & 0.6831 & 0.7177 & 0.7218 & 0.6896 & 0.6291 & 30.49\\
    \hline
    \textbf{Mid checkpoint by FID}  & 0.7005 & 0.7241 & \textbf{0.7374} & 0.7115 & 0.6270 & 37.23\\
    \hline
    \textbf{Worst checkpoint by FID}  & 0.6775 & 0.7184 & 0.7229 & 0.6931 & 0.6218 & 72.48\\
    \hline
\end{tabular}
\caption{Tile-and-Stitch - 2D}
\label{tab:Tile-and-Stitch - 2D}
\end{table*}

\begin{table*}[ht]
\centering
\setlength{\tabcolsep}{2pt}
\begin{tabular}{
  | >{\raggedright\arraybackslash}p{4cm}
  | R{1.45cm} 
  | R{1.45cm} 
  | R{1.44cm} 
  | R{1.58cm} 
  | R{1.58cm}
  | R{1cm} |
}

\hline
\textbf{Unet-models: \textit{Loss, Metric}} 
& \multicolumn{1}{c|}{\textit{Dice, IoU}} 
& \multicolumn{1}{c|}{\textit{MSE, IoU}} 
& \multicolumn{1}{c|}{\textit{BCE, IoU}} 
& \multicolumn{1}{c|}{\textit{Dice, MSE}} 
& \multicolumn{1}{c|}{\textit{MSE, Dice}} 
& \multicolumn{1}{c|}{\textbf{FID}}\\
    \hline
    \textbf{Best checkpoint by FID}  & 0.6478 & 0.6910 & 0.6906 & 0.6685 & 0.5943 & 40.74\\
    \hline
    \textbf{Mid checkpoint by FID}  & 0.6496 & 0.6901 & \textbf{0.6927} & 0.6696 & 0.5821 & 42.53\\
    \hline
    \textbf{Worst checkpoint by FID}  & 0.6486 & 0.6739 & 0.6860 & 0.6637 & 0.5852 & 53.23\\
    \hline
\end{tabular}
\caption{Tile-and-Stitch - 2D (orthoslice)}
\label{tab:Tile-and-Stitch - 2D (orthoslice)}
\end{table*}

\begin{table*}[ht]
\centering
\setlength{\tabcolsep}{2pt}
\begin{tabular}{
  | >{\raggedright\arraybackslash}p{4cm}
  | R{1.45cm} 
  | R{1.45cm} 
  | R{1.44cm} 
  | R{1.58cm} 
  | R{1.58cm}
  | R{1cm} |
}

\hline
\textbf{Unet-models: \textit{Loss, Metric}} 
& \multicolumn{1}{c|}{\textit{Dice, IoU}} 
& \multicolumn{1}{c|}{\textit{MSE, IoU}} 
& \multicolumn{1}{c|}{\textit{BCE, IoU}} 
& \multicolumn{1}{c|}{\textit{Dice, MSE}} 
& \multicolumn{1}{c|}{\textit{MSE, Dice}} 
& \multicolumn{1}{c|}{\textbf{FID}}\\
    \hline
    \textbf{Best checkpoint by FID}  & 0.6867 & 0.6873 & 0.7155 & 0.6812 & 0.6075 & 36.99\\
    \hline
    \textbf{Mid checkpoint by FID}  & 0.6998 & 0.7202 & \textbf{0.7359} & 0.7127 & 0.6167 & 40.91\\
    \hline
    \textbf{Worst checkpoint by FID}  & 0.5870 & 0.5218 & 0.6555 & 0.6429 & 0.5476 & 120.13\\
    \hline
\end{tabular}
\caption{Valid-padding, no crop - 3D}
\label{tab:Valid-padding, no crop - 3D}
\end{table*}

\begin{table*}[ht]
\centering
\setlength{\tabcolsep}{2pt}
\begin{tabular}{
  | >{\raggedright\arraybackslash}p{4cm}
  | R{1.45cm} 
  | R{1.45cm} 
  | R{1.44cm} 
  | R{1.58cm} 
  | R{1.58cm}
  | R{1cm} |
}

\hline
\textbf{Unet-models: \textit{Loss, Metric}} 
& \multicolumn{1}{c|}{\textit{Dice, IoU}} 
& \multicolumn{1}{c|}{\textit{MSE, IoU}} 
& \multicolumn{1}{c|}{\textit{BCE, IoU}} 
& \multicolumn{1}{c|}{\textit{Dice, MSE}} 
& \multicolumn{1}{c|}{\textit{MSE, Dice}} 
& \multicolumn{1}{c|}{\textbf{FID}}\\
    \hline
    \textbf{Best checkpoint by FID}  & 0.6828 & 0.7163 & \textbf{0.7230} & 0.6908 & 0.6286 & 30.23\\
    \hline
    \textbf{Mid checkpoint by FID}  & 0.6796 & 0.7124 & 0.7215 & 0.6876 & 0.6155 & 34.78\\
    \hline
    \textbf{Worst checkpoint by FID}  & 0.6681 & 0.7075 & 0.7149 & 0.6784 & 0.6153 & 64.62\\
    \hline
\end{tabular}
\caption{Valid-padding, no crop - 2D}
\label{tab:Valid-padding, no crop - 2D}
\end{table*}

\begin{table*}[ht]
\centering
\setlength{\tabcolsep}{2pt}
\begin{tabular}{
  | >{\raggedright\arraybackslash}p{4cm}
  | R{1.45cm} 
  | R{1.45cm} 
  | R{1.44cm} 
  | R{1.58cm} 
  | R{1.58cm}
  | R{1cm} |
}

\hline
\textbf{Unet-models: \textit{Loss, Metric}} 
& \multicolumn{1}{c|}{\textit{Dice, IoU}} 
& \multicolumn{1}{c|}{\textit{MSE, IoU}} 
& \multicolumn{1}{c|}{\textit{BCE, IoU}} 
& \multicolumn{1}{c|}{\textit{Dice, MSE}} 
& \multicolumn{1}{c|}{\textit{MSE, Dice}} 
& \multicolumn{1}{c|}{\textbf{FID}}\\
    \hline
    \textbf{Best checkpoint by FID}  & 0.6832 & 0.7235 & 0.7304 & 0.6996 & 0.6250 & 40.71\\
    \hline
    \textbf{Mid checkpoint by FID}  & 0.6886 & 0.7219 & \textbf{0.7334} & 0.7040 & 0.6202 & 42.53\\
    \hline
    \textbf{Worst checkpoint by FID}  & 0.6649 & 0.7113 & 0.7184 & 0.6765 & 0.6130 & 50.28\\
    \hline
\end{tabular}
\caption{Valid-padding (no crop) - 2D (orthoslice)}
\label{tab:Valid-padding (no crop) - 2D (orthoslice)}
\end{table*}

\begin{table*}[ht]
\centering
\setlength{\tabcolsep}{2pt}
\begin{tabular}{
  | >{\raggedright\arraybackslash}p{4cm}
  | R{1.45cm} 
  | R{1.45cm} 
  | R{1.44cm} 
  | R{1.58cm} 
  | R{1.58cm}
  | R{1cm} |
}

\hline
\textbf{Unet-models: \textit{Loss, Metric}} 
& \multicolumn{1}{c|}{\textit{Dice, IoU}} 
& \multicolumn{1}{c|}{\textit{MSE, IoU}} 
& \multicolumn{1}{c|}{\textit{BCE, IoU}} 
& \multicolumn{1}{c|}{\textit{Dice, MSE}} 
& \multicolumn{1}{c|}{\textit{MSE, Dice}} 
& \multicolumn{1}{c|}{\textbf{FID}}\\
    \hline
    \textbf{Best checkpoint by FID}  & 0.6780 & 0.6880 & 0.7120 & 0.6807 & 0.6184 & 41.21\\
    \hline
    \textbf{Mid checkpoint by FID}  & 0.7014 & 0.7197 & \textbf{0.7382} & 0.7153 & 0.6187 & 47.85\\
    \hline
    \textbf{Worst checkpoint by FID}  & 0.5788 & 0.4127 & 0.6923 & 0.6699 & 0.5855 & 188.23\\
    \hline
\end{tabular}
\caption{Same-padding - 3D}
\label{tab:Same-padding - 3D}
\end{table*}

\begin{table*}[ht]
\centering
\setlength{\tabcolsep}{2pt}
\begin{tabular}{
  | >{\raggedright\arraybackslash}p{4cm}
  | R{1.45cm} 
  | R{1.45cm} 
  | R{1.44cm} 
  | R{1.58cm} 
  | R{1.58cm}
  | R{1cm} |
}

\hline
\textbf{Unet-models: \textit{Loss, Metric}} 
& \multicolumn{1}{c|}{\textit{Dice, IoU}} 
& \multicolumn{1}{c|}{\textit{MSE, IoU}} 
& \multicolumn{1}{c|}{\textit{BCE, IoU}} 
& \multicolumn{1}{c|}{\textit{Dice, MSE}} 
& \multicolumn{1}{c|}{\textit{MSE, Dice}} 
& \multicolumn{1}{c|}{\textbf{FID}}\\
    \hline
    \textbf{Best checkpoint by FID}  & 0.6831 & 0.7178 & 0.7220 & 0.6896 & 0.6288 & 30.45\\
    \hline
    \textbf{Mid checkpoint by FID}  & 0.6999 & 0.7235 & \textbf{0.7372} & 0.7110 & 0.6276 & 37.07\\
    \hline
    \textbf{Worst checkpoint by FID}  & 0.6769 & 0.7173 & 0.7219 & 0.6919 & 0.6197 & 71.76\\
    \hline
\end{tabular}
\caption{Same-padding - 2D}
\label{tab:Same-padding - 2D}
\end{table*}

\begin{table}
\centering
\setlength{\tabcolsep}{2pt}
\begin{tabular}{
  | >{\raggedright\arraybackslash}p{4cm}
  | R{1.45cm} 
  | R{1.45cm} 
  | R{1.44cm} 
  | R{1.58cm} 
  | R{1.58cm}
  | R{1cm} |
}

\hline
\textbf{Unet-models: \textit{Loss, Metric}} 
& \multicolumn{1}{c|}{\textit{Dice, IoU}} 
& \multicolumn{1}{c|}{\textit{MSE, IoU}} 
& \multicolumn{1}{c|}{\textit{BCE, IoU}} 
& \multicolumn{1}{c|}{\textit{Dice, MSE}} 
& \multicolumn{1}{c|}{\textit{MSE, Dice}} 
& \multicolumn{1}{c|}{\textbf{FID}}\\
    \hline
    \textbf{Best checkpoint by FID}  & 0.6826 & 0.7257 & 0.7308 & 0.7003 & 0.6245 & 41.02\\
    \hline
    \textbf{Mid checkpoint by FID}  & 0.6863 & 0.7235 & \textbf{0.7325} & 0.7025 & 0.6146 & 42.50\\
    \hline
    \textbf{Worst checkpoint by FID}  & 0.6825 & 0.7059 & 0.7245 & 0.6923 & 0.6148 & 53.48\\
    \hline
\end{tabular}
\caption{Same-padding - 2D (orthoslice)}
\label{tab:Same-padding - 2D (orthoslice)}
\end{table}

\subsection{Visual evaluation of segmentation results from different dimensionalities}
 Here, we show an example of the results from the supplementary tables for the case of the tile-and-stitch method, using just one of the Unet models (with Dice Loss and IoU evaluation metric).

\subsection{Effect of hyperparameter choice on FID, for style-transfer in EM dataset sampled from openorganelle database.}

\printbibliography

@article{lauenburg2022instance,
  title={Instance segmentation of unlabeled modalities via cyclic segmentation gan},
  author={Lauenburg, Leander and Lin, Zudi and Zhang, Ruihan and Santos, M{\'a}rcia dos and Huang, Siyu and Arganda-Carreras, Ignacio and Boyden, Edward S and Pfister, Hanspeter and Wei, Donglai},
  journal={arXiv preprint arXiv:2204.03082},
  year={2022}
}

@inproceedings{rumberger2021shift,
  title={How shift equivariance impacts metric learning for instance segmentation},
  author={Rumberger, Josef Lorenz and Yu, Xiaoyan and Hirsch, Peter and Dohmen, Melanie and Guarino, Vanessa Emanuela and Mokarian, Ashkan and Mais, Lisa and Funke, Jan and Kainmueller, Dagmar},
  booktitle={Proceedings of the IEEE/CVF International Conference on Computer Vision},
  pages={7128--7136},
  year={2021}
}

@inproceedings{wei2020mitoem,
  title={Mitoem dataset: Large-scale 3d mitochondria instance segmentation from em images},
  author={Wei, Donglai and Lin, Zudi and Franco-Barranco, Daniel and Wendt, Nils and Liu, Xingyu and Yin, Wenjie and Huang, Xin and Gupta, Aarush and Jang, Won-Dong and Wang, Xueying and others},
  booktitle={International Conference on Medical Image Computing and Computer-Assisted Intervention},
  pages={66--76},
  year={2020},
  organization={Springer}
}

@inproceedings{isola2017image,
  title={Image-to-image translation with conditional adversarial networks},
  author={Isola, Phillip and Zhu, Jun-Yan and Zhou, Tinghui and Efros, Alexei A},
  booktitle={Proceedings of the IEEE conference on computer vision and pattern recognition},
  pages={1125--1134},
  year={2017}
}

@inproceedings{zhu2017unpaired,
  title={Unpaired image-to-image translation using cycle-consistent adversarial networks},
  author={Zhu, Jun-Yan and Park, Taesung and Isola, Phillip and Efros, Alexei A},
  booktitle={Proceedings of the IEEE international conference on computer vision},
  pages={2223--2232},
  year={2017}
}

@inproceedings{he2016deep,
  title={Deep residual learning for image recognition},
  author={He, Kaiming and Zhang, Xiangyu and Ren, Shaoqing and Sun, Jian},
  booktitle={Proceedings of the IEEE conference on computer vision and pattern recognition},
  pages={770--778},
  year={2016}
}

@inproceedings{zhang2018translating,
  title={Translating and segmenting multimodal medical volumes with cycle-and shape-consistency generative adversarial network},
  author={Zhang, Zizhao and Yang, Lin and Zheng, Yefeng},
  booktitle={Proceedings of the IEEE conference on computer vision and pattern Recognition},
  pages={9242--9251},
  year={2018}
}

@article{de2021residual,
  title={Residual cyclegan for robust domain transformation of histopathological tissue slides},
  author={de Bel, Thomas and Bokhorst, John-Melle and van der Laak, Jeroen and Litjens, Geert},
  journal={Medical Image Analysis},
  volume={70},
  pages={102004},
  year={2021},
  publisher={Elsevier}
}

@article{saalfeld2019computational,
  title={Computational methods for stitching, alignment, and artifact correction of serial section data},
  author={Saalfeld, Stephan},
  journal={Methods in Cell Biology},
  volume={152},
  pages={261--276},
  year={2019},
  publisher={Elsevier}
}

@inproceedings{cciccek20163d,
  title={3D U-Net: learning dense volumetric segmentation from sparse annotation},
  author={{\c{C}}i{\c{c}}ek, {\"O}zg{\"u}n and Abdulkadir, Ahmed and Lienkamp, Soeren S and Brox, Thomas and Ronneberger, Olaf},
  booktitle={International conference on medical image computing and computer-assisted intervention},
  pages={424--432},
  year={2016},
  organization={Springer}
}

@inproceedings{kayhan2020translation,
  title={On translation invariance in cnns: Convolutional layers can exploit absolute spatial location},
  author={Kayhan, Osman Semih and Gemert, Jan C van},
  booktitle={Proceedings of the IEEE/CVF conference on computer vision and pattern recognition},
  pages={14274--14285},
  year={2020}
}

@article{preibisch2009globally,
  title={Globally optimal stitching of tiled 3D microscopic image acquisitions},
  author={Preibisch, Stephan and Saalfeld, Stephan and Tomancak, Pavel},
  journal={Bioinformatics},
  volume={25},
  number={11},
  pages={1463--1465},
  year={2009},
  publisher={Oxford University Press}
}

@inproceedings{ronneberger2015u,
  title={U-net: Convolutional networks for biomedical image segmentation},
  author={Ronneberger, Olaf and Fischer, Philipp and Brox, Thomas},
  booktitle={International Conference on Medical image computing and computer-assisted intervention},
  pages={234--241},
  year={2015},
  organization={Springer}
}

@article{reina2020systematic,
  title={Systematic evaluation of image tiling adverse effects on deep learning semantic segmentation},
  author={Reina, G Anthony and Panchumarthy, Ravi and Thakur, Siddhesh Pravin and Bastidas, Alexei and Bakas, Spyridon},
  journal={Frontiers in neuroscience},
  volume={14},
  pages={65},
  year={2020},
  publisher={Frontiers Media SA}
}

@article{possolo2021exact,
  title={Exact tile-based segmentation inference for images larger than gpu memory},
  author={Possolo, Michael and Bajcsy, Peter},
  journal={Journal of Research of the National Institute of Standards and Technology},
  volume={126},
  pages={126009},
  year={2021}
}

@article{buglakova2025tiling,
  title={Tiling artifacts and trade-offs of feature normalization in the segmentation of large biological images},
  author={Buglakova, Elena and Archit, Anwai and D'Imprima, Edoardo and Mahamid, Julia and Pape, Constantin and Kreshuk, Anna},
  journal={arXiv preprint arXiv:2503.19545},
  year={2025}
}

@article{bria2012terastitcher,
  title={TeraStitcher-a tool for fast automatic 3D-stitching of teravoxel-sized microscopy images},
  author={Bria, Alessandro and Iannello, Giulio},
  journal={BMC bioinformatics},
  volume={13},
  number={1},
  pages={316},
  year={2012},
  publisher={Springer}
}

@article{heinrich2021whole,
  title={Whole-cell organelle segmentation in volume electron microscopy},
  author={Heinrich, Larissa and Bennett, Davis and Ackerman, David and Park, Woohyun and Bogovic, John and Eckstein, Nils and Petruncio, Alyson and Clements, Jody and Pang, Song and Xu, C Shan and others},
  journal={Nature},
  volume={599},
  number={7883},
  pages={141--146},
  year={2021},
  publisher={Nature Publishing Group UK London}
}

@article{Xu2020,
author = "C. Shan Xu and Gleb Shtengel and Davis Bennett and Aubrey Weigel and Amalia Pasolli and Harald Hess",
title = "{Isotropic 3D electron microscopy reference data of wild-type, interphase HeLa cell (jrc_hela-1)}",
year = "2020",
month = "11",
url = "https://janelia.figshare.com/articles/dataset/Isotropic_3D_electron_microscopy_reference_data_of_wild-type_interphase_HeLa_cell_jrc_hela-1_/13123415",
doi = "10.25378/janelia.13123415.v1"
}

@article{jrcHela-2,
author = "FIB-SEM Technology Group and CellMap Project Team and Davis Bennett and Harald Hess and Amalia Pasolli and Gleb Shtengel and Aubrey Weigel and C. Shan Xu",
title = "{Isotropic 3D electron microscopy reference data of wild-type, interphase HeLa cell (jrc_hela-2)}",
year = "2020",
month = "11",
url = "https://janelia.figshare.com/articles/dataset/Isotropic_3D_electron_microscopy_reference_data_of_wild-type_interphase_HeLa_cell_jrc_hela-2_/13114211",
doi = "10.25378/janelia.13114211.v2"
}

@article{jrcHela-3,
author = "FIB-SEM Technology Group and CellMap Project Team and Davis Bennett and Harald Hess and Amalia Pasolli and Gleb Shtengel and Aubrey Weigel and C. Shan Xu",
title = "{Isotropic 3D electron microscopy reference data of wild-type, interphase HeLa cell (jrc_hela-3)}",
year = "2020",
month = "11",
url = "https://janelia.figshare.com/articles/dataset/Isotropic_3D_electron_microscopy_reference_data_of_wild-type_interphase_HeLa_cell_jrc_hela-3_/13114244",
doi = "10.25378/janelia.13114244.v2"
}

@article{jrcjurkat-1,
author = "FIB-SEM Technology Group and CellMap Project Team and Davis Bennett and Harald Hess and Huxley Hoffman and Schuyler van Engelenburg and Gleb Shtengel and C. Shan Xu",
title = "{Isotropic 3D electron microscopy reference data of wild-type, immortalized T-Cells (jrc_jurkat-1)}",
year = "2020",
month = "11",
url = "https://janelia.figshare.com/articles/dataset/Isotropic_3D_electron_microscopy_reference_data_of_wild-type_immortalized_T-Cells_jrc_jurkat-1_/13114259",
doi = "10.25378/janelia.13114259.v2"
}

@misc{Seitzer2020FID,
  author={Maximilian Seitzer},
  title={{pytorch-fid: FID Score for PyTorch}},
  month=8,
  year={2020},
  note={Version 0.3.0},
  howpublished={\url{https://github.com/mseitzer/pytorch-fid}},
}

@article{heusel2017gans,
  title={Gans trained by a two time-scale update rule converge to a local nash equilibrium},
  author={Heusel, Martin and Ramsauer, Hubert and Unterthiner, Thomas and Nessler, Bernhard and Hochreiter, Sepp},
  journal={Advances in neural information processing systems},
  volume={30},
  year={2017}
}

@article{jansche2025deep,
  title={Deep learning-based image super resolution methods in microscopy--a review},
  author={Jansche, Andreas and Krawczyk, Patrick and Balaguera, Miguelangel and Kini, Anoop and Bernthaler, Timo and Schneider, Gerhard},
  journal={Methods in Microscopy},
  volume={2},
  number={2},
  pages={235--275},
  year={2025},
  publisher={De Gruyter}
}

@article{eschweiler20213d,
  title={3D fluorescence microscopy data synthesis for segmentation and benchmarking},
  author={Eschweiler, Dennis and Rethwisch, Malte and Jarchow, Mareike and Koppers, Simon and Stegmaier, Johannes},
  journal={Plos one},
  volume={16},
  number={12},
  pages={e0260509},
  year={2021},
  publisher={Public Library of Science San Francisco, CA USA}
}

@article{thambawita2022singan,
  title={SinGAN-Seg: Synthetic training data generation for medical image segmentation},
  author={Thambawita, Vajira and Salehi, Pegah and Sheshkal, Sajad Amouei and Hicks, Steven A and Hammer, Hugo L and Parasa, Sravanthi and Lange, Thomas de and Halvorsen, P{\aa}l and Riegler, Michael A},
  journal={PloS one},
  volume={17},
  number={5},
  pages={e0267976},
  year={2022},
  publisher={Public Library of Science San Francisco, CA USA}
}

@article{khader2023denoising,
  title={Denoising diffusion probabilistic models for 3D medical image generation},
  author={Khader, Firas and M{\"u}ller-Franzes, Gustav and Tayebi Arasteh, Soroosh and Han, Tianyu and Haarburger, Christoph and Schulze-Hagen, Maximilian and Schad, Philipp and Engelhardt, Sandy and Bae{\ss}ler, Bettina and Foersch, Sebastian and others},
  journal={Scientific Reports},
  volume={13},
  number={1},
  pages={7303},
  year={2023},
  publisher={Nature Publishing Group UK London}
}

@article{demiray2021d,
  title={D-SRGAN: DEM super-resolution with generative adversarial networks},
  author={Demiray, Bekir Z and Sit, Muhammed and Demir, Ibrahim},
  journal={SN Computer Science},
  volume={2},
  number={1},
  pages={48},
  year={2021},
  publisher={Springer}
}

@article{saharia2022image,
  title={Image super-resolution via iterative refinement},
  author={Saharia, Chitwan and Ho, Jonathan and Chan, William and Salimans, Tim and Fleet, David J and Norouzi, Mohammad},
  journal={IEEE transactions on pattern analysis and machine intelligence},
  volume={45},
  number={4},
  pages={4713--4726},
  year={2022},
  publisher={IEEE}
}

@inproceedings{ledig2017photo,
  title={Photo-realistic single image super-resolution using a generative adversarial network},
  author={Ledig, Christian and Theis, Lucas and Husz{\'a}r, Ferenc and Caballero, Jose and Cunningham, Andrew and Acosta, Alejandro and Aitken, Andrew and Tejani, Alykhan and Totz, Johannes and Wang, Zehan and others},
  booktitle={Proceedings of the IEEE conference on computer vision and pattern recognition},
  pages={4681--4690},
  year={2017}
}

@inproceedings{liu2015faceattributes,
  title = {Deep Learning Face Attributes in the Wild},
  author = {Liu, Ziwei and Luo, Ping and Wang, Xiaogang and Tang, Xiaoou},
  booktitle = {Proceedings of International Conference on Computer Vision (ICCV)},
  month = 12,
  year = {2015} 
}

@article{ILSVRC15,
Author = {Olga Russakovsky and Jia Deng and Hao Su and Jonathan Krause and Sanjeev Satheesh and Sean Ma and Zhiheng Huang and Andrej Karpathy and Aditya Khosla and Michael Bernstein and Alexander C. Berg and Li Fei-Fei},
Title = {{ImageNet Large Scale Visual Recognition Challenge}},
Year = {2015},
journal   = {International Journal of Computer Vision (IJCV)},
doi = {10.1007/s11263-015-0816-y},
volume={115},
number={3},
pages={211-252}
}

@inproceedings{yu2014fine,
  title={Fine-grained visual comparisons with local learning},
  author={Yu, Aron and Grauman, Kristen},
  booktitle={Proceedings of the IEEE conference on computer vision and pattern recognition},
  pages={192--199},
  year={2014}
}

@inproceedings{yu2017semantic,
  title={Semantic jitter: Dense supervision for visual comparisons via synthetic images},
  author={Yu, Aron and Grauman, Kristen},
  booktitle={Proceedings of the IEEE International Conference on Computer Vision},
  pages={5570--5579},
  year={2017}
}

@article{kieselmann2021cross,
  title={Cross-modality deep learning: contouring of MRI data from annotated CT data only},
  author={Kieselmann, Jennifer P and Fuller, Clifton D and Gurney-Champion, Oliver J and Oelfke, Uwe},
  journal={Medical physics},
  volume={48},
  number={4},
  pages={1673--1684},
  year={2021},
  publisher={Wiley Online Library}
}

@article{jin2017cyclegan,
  title={Cyclegan face-off},
  author={Jin, Xiaohan and Qi, Ye and Wu, Shangxuan},
  journal={arXiv preprint arXiv:1712.03451},
  year={2017}
}

@article {wolny2020accurate,
article_type = {journal},
title = {Accurate and versatile 3D segmentation of plant tissues at cellular resolution},
author = {Wolny, Adrian and Cerrone, Lorenzo and Vijayan, Athul and Tofanelli, Rachele and Barro, Amaya Vilches and Louveaux, Marion and Wenzl, Christian and Strauss, Sören and Wilson-Sánchez, David and Lymbouridou, Rena and Steigleder, Susanne S and Pape, Constantin and Bailoni, Alberto and Duran-Nebreda, Salva and Bassel, George W and Lohmann, Jan U and Tsiantis, Miltos and Hamprecht, Fred A and Schneitz, Kay and Maizel, Alexis and Kreshuk, Anna},
editor = {Hardtke, Christian S and Bergmann, Dominique C and Bergmann, Dominique C and Graeff, Moritz},
volume = 9,
year = 2020,
month = 7,
pub_date = {2020-07-29},
pages = {e57613},
citation = {eLife 2020;9:e57613},
doi = {10.7554/eLife.57613},
url = {https://doi.org/10.7554/eLife.57613},
journal = {eLife},
issn = {2050-084X},
publisher = {eLife Sciences Publications, Ltd},
}

\end{document}